\documentclass[letterpaper, 10 pt, conference]{ieeeconf}

\IEEEoverridecommandlockouts

\usepackage{multicol}
\usepackage{bbm}
\usepackage{algorithm}
\usepackage{algpseudocode}
\usepackage{graphicx}
\usepackage{cuted}
\usepackage{caption}
\usepackage{subcaption}
\usepackage{placeins}
\usepackage{float}
\usepackage{stfloats}
\usepackage{amsmath}
\usepackage{amssymb}
\usepackage{xcolor, colortbl}
\usepackage{hyperref}
\usepackage{multirow}
\usepackage{capt-of}
\usepackage{tabularx}
\usepackage{booktabs}
\usepackage{tabularray}
\usepackage{threeparttable}
\newcommand{\pr}[1]{\textbf{ #1}}

\title{\LARGE
\textit{Risk-Guided Diffusion}: Toward Deploying Robot Foundation Models In Space, Where Failure Is Not An Option
}

\author{
$^*$Rohan Thakker$^1$,
$^*$Adarsh Patnaik$^{1,3}$,
Vince Kurtz$^2$,
Jonas Frey$^3$,
Jonathan Becktor$^1$,
Sangwoo Moon$^1$,\\
Rob Royce$^1$,
Marcel Kaufmann$^1$,
Georgios Georgakis$^1$,
Pascal Roth$^3$,\\
Joel Burdick$^2$,
Marco Hutter$^3$,
Shehryar Khattak$^1$\\
$^1$\textit{NASA-JPL, Caltech} \quad
$^2$\textit{CME, Caltech} \quad
$^3$\textit{Robotic Systems Lab, ETH Zurich} \quad
\thanks{The research was carried out at the Jet Propulsion Laboratory, California Institute of Technology, under a contract with the National Aeronautics and Space Administration (80NM0018D0004).
\copyright 2024. California Institute of Technology. Government sponsorship acknowledged. All rights reserved.}
}

\begin{document}
\maketitle
\begin{strip}
\vskip-75pt
\centering
\includegraphics[width=0.9\textwidth, page=4, trim=0cm 26.5cm 0cm 0cm, clip]{figs/system12_v9.pdf}\\
\noindent\hspace*{\fill}
\begin{minipage}[t]{0.25\linewidth}
\centering
\vspace{-4.4cm}
\includegraphics[width=0.8\linewidth, page=2, trim=0cm 22cm 16cm 0cm, clip]{figs/system12_v9.pdf}\\
\vspace{0.6cm}
\small (a) Rover drives away from potential evidence of water due to lack of semantic awareness.
\end{minipage}
\begin{minipage}[t]{0.35\linewidth}
\centering
\includegraphics[width=0.99\linewidth, page=1, trim=0cm 21cm 12cm 0cm, clip]{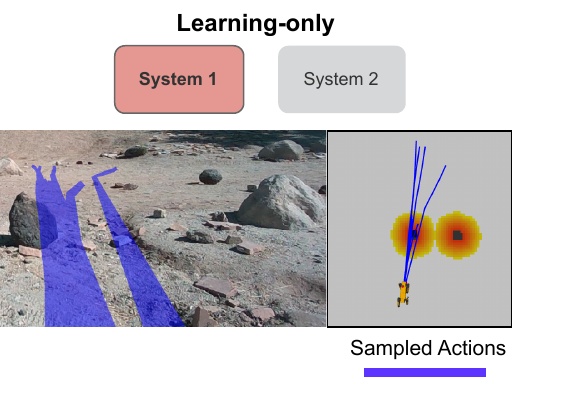}\\
\small (b) Navigation foundation model fails to identify a hazard not seen in training data.
\end{minipage}
\begin{minipage}[t]{0.35\linewidth}
\centering
\includegraphics[width=0.98\linewidth, page=3, trim=0cm 21cm 12cm 0cm, clip]{figs/system12_v9.pdf}\\
\small (c) Robot successfully avoids hazard while following semantic image-based goals.
\end{minipage}
\captionof{figure}{
An example robotic planetary surface exploration mission scenario to provide motivation for the proposed hybrid learning and physics-based approaches. (a) shows ignoring semantic information completely may be detrimental towards overall mission success, (b) shows failure of learning-based approaches to adapt to out-of-distribution hazards during runtime, (c) shows viability of the proposed hybrid approaches.
}
\label{fig:system12}
\vspace{-3ex}
\end{strip}

\begin{abstract}
Safe, reliable navigation in extreme, unfamiliar terrain is required for future robotic space exploration missions.
Recent generative-AI methods learn semantically aware navigation policies from large, cross-embodiment datasets, but offer limited safety guarantees.
Inspired by human cognitive science \cite{daniel2017thinking}, we propose a risk-guided diffusion framework that fuses a fast, learned ``System-1" with a slow, physics-based ``System-2," sharing computation at both training and inference to couple adaptability with formal safety.
Hardware experiments conducted at the NASA JPL's Mars-analog facility, Mars Yard, show that our approach reduces failure rates by up to $4\times$ while matching the goal-reaching performance of learning-based robotic models by leveraging inference-time compute without any additional training.
\end{abstract}

\section{Introduction}
NASA's surface exploration missions currently rely on pre-mission maps and ground supervision. This limits robotic exploration of distant worlds like Enceladus or Europa, where one-way communication latency nears 50 minutes and outages can persist for weeks \cite{ono2024boldly, thakker2024boldly}.
Future robots need to autonomously traverse GNSS-denied, visually degraded terrain with no prior maps to successfully conduct their missions.
Robotic foundation models trained on cross-embodiment datasets have demonstrated strong vision-language reasoning \cite{kim2024openvla, collaboration2023open, shah2023gnm}, but their reliability drops for out-of-distribution scenarios.
This work asks: \textbf{Can foundation models deliver adaptive autonomous mobility in space environments without compromising safety?}

Adding safety to learning-based navigation methods has seen considerable work over the years.
Prior works like \cite{dixit2023step, freyfast} have shown the benefits of using a traversability estimation module to plan safe paths.
Several foundation models have benefited from test time compute scaling \cite{setlur2025scaling}.
Works such as \cite{Liang2024DTGD} use a binary traversability mask at training time and modify the loss function of a diffusion model to output trajectories in traversable regions.
SafeDiffuser \cite{xiao2023safediffuser} provides safety guarantees using a Control Barrier Function (CBF) \cite{ames2016control} at inference time, but does not account for stochasticity in traversability estimation.
Furthermore, CBF construction can be challenging for stochastic systems.

Inspired by Kahneman \cite{daniel2017thinking}, several recent works have been proposed to combine slow and fast thinking \cite{bjorck2025gr00t,figure2024helix,yoon2025monte}.
In this work, we explore the possibility of using a learning based navigation foundation model as \textit{fast} System 1 and physics-based risk-assessment as \textit{slow} System 2.
\pr{Contributions:} (1) We enable \textbf{safe adaptation} to out-of-training-distribution terrains by leveraging stochastic physics-based traversability estimation \cite{dixit2023step}.
(2) \cite{sridhar2024nomad, chi2024diffusionpolicy} achieve \textbf{cross-embodied transfer} by abstracting the action space to a sequence of waypoints with an underlying low-level tracking controller.
However, every robot has a different size and traversability capability.
To address this, we leverage a diffusion-based action head to learn a multi-modal action distribution that represent different homotopy classes of candidate trajectories.
(3) We leverage \textbf{inference-time compute} with physics-based models to improve performance without additional training data.
Using an intuitive 1-D example, we show that standard classifier guidance \cite{dhariwal2021diffusion} is insufficient for constraint satisfaction and propose an extension of a projection-based guidance strategy \cite{christopher2024constrained}.
(4) \textbf{Experimental validation in NASA JPL's Mars-analog facility} shows our approach reduces failure rates up to $4\times$ without reducing the goal-reaching performance of learning-based models.

\section{Approach}

\begin{figure*}[t!]
    \centering
    \includegraphics[width=\linewidth]{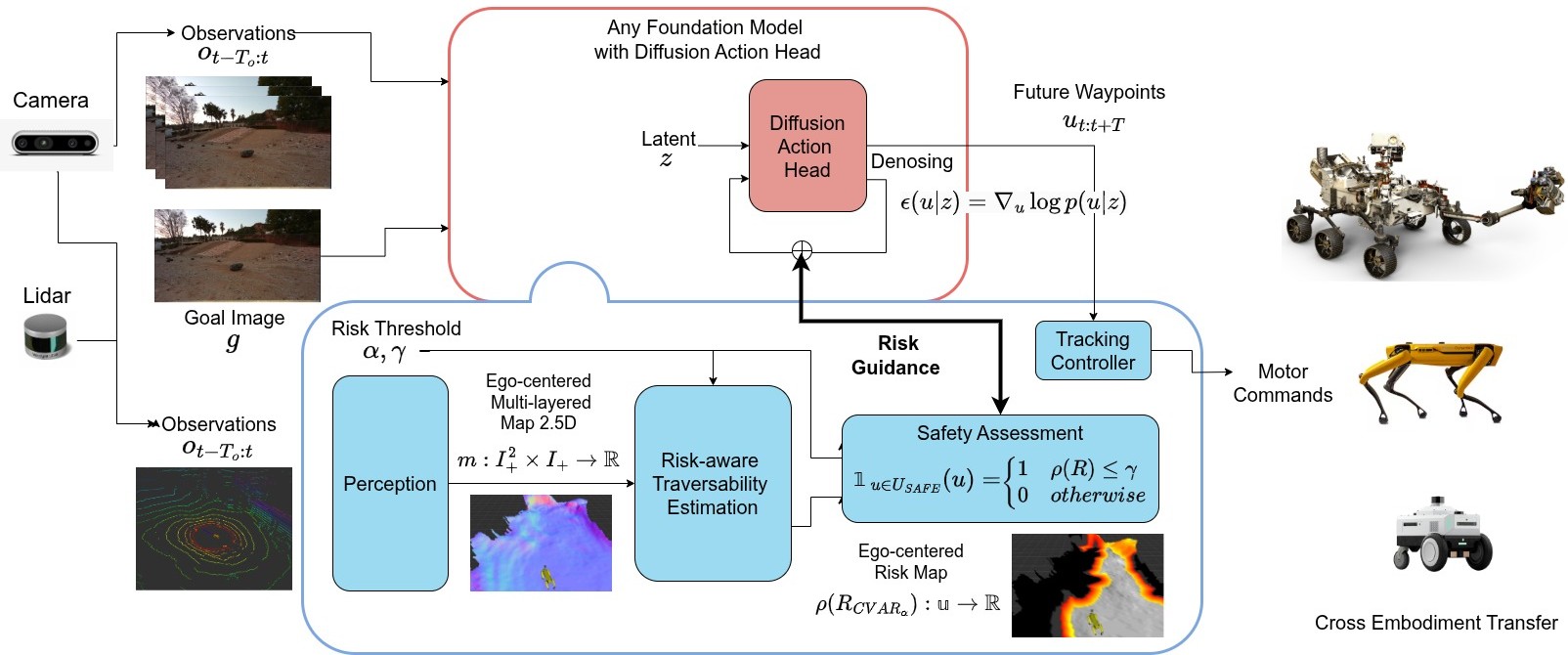}
    \caption{
    Risk-Guided Diffusion Architecture Overview.
    }
    \label{fig:architecture-overview}
    \vspace{-3ex}
\end{figure*}

\begin{figure*}[h!]
   \centering

   \begin{subfigure}[b]{0.65\textwidth}
       \centering
       \includegraphics[width=\linewidth, page=2, trim=0.5cm 4.5cm 1cm 4.5cm, clip]{figs/1d_example.pdf}
       \caption{Diffusion model without any system 2 guidance.}
       \label{fig:1d_no_guidance}
   \end{subfigure}
   \begin{subfigure}[b]{0.62\textwidth}
       \centering
       \includegraphics[width=\linewidth, page=1, trim=3cm 6cm 3cm 7.0cm, clip]{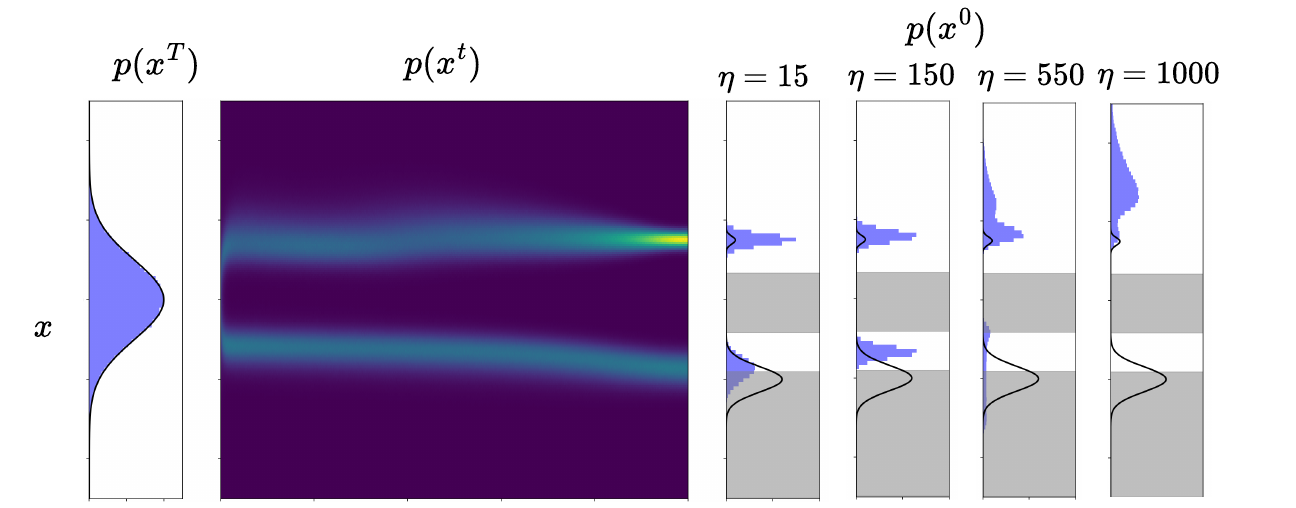}
       \caption{Classifier-based guidance \cite{dhariwal2021diffusion} with varying penalty weights $\eta$}
       \label{fig:1d_classifier}
   \end{subfigure}
   \begin{subfigure}[b]{0.37\textwidth}
       \centering
       \includegraphics[width=\linewidth, page=3, trim=2.3cm 6cm 12cm 7.5cm, clip]{figs/1d_example.pdf}
       \caption{Projection-based guidance \cite{christopher2024constrained}}
       \label{fig:1d_projection}
   \end{subfigure}

   \caption{Langevin sampling dynamics for different guidance strategies in 1D rover navigation.}
   \label{fig:1d_example}
   \vspace{-3ex}
\end{figure*}

Inspired by Kahneman \cite{daniel2017thinking}, this section presents a framework that combines a fast, intuitive ``System 1" (pre-trained foundation model) with a slow, deliberate ``System 2" (physics-based risk estimator).

\subsection{System 1: Learning-based Foundation Model}

Mathematically, System 1 is a pre-trained control policy $\pi(u|o,g)$ that outputs $N_u$ waypoints $u_{N_u} \in \mathbb{R}^{2 \times N_u}$ given $N_o$ image observations $o_{N_o}$ and goal image $o_g$.
Following \cite{sridhar2024nomad}, we encode high-dimensional observations in a latent representation $z$.
The policy uses a latent diffusion model $\pi(u|z)$ defined by the score function $\epsilon_\theta(u^t, t|z)$, which is trained with standard techniques \cite{song2020score},
\begin{gather}
    \min_{\theta} \mathbb{E}_{u, z \sim \mathcal{D}, t\sim Uniform(0,T)}
    [||\epsilon - \epsilon_\theta(u^t, t|z)||^2], \\
    u^t = \sigma^t u^0 + \alpha^t \epsilon, \quad \epsilon \sim N(0,I),
\end{gather}
where $u^T \sim \mathcal{N}(0, I)$, $u^0 \sim \pi(u|z)$, and $\alpha^t, \sigma^t$ are noise schedule parameters. We use $(.)^{t_{diffusion}}_{t_{trajectory}}$ notation for diffusion vs. trajectory time.

This system has three key features: (1) it uses train-time compute to learn a foundation model that excels at intuitive reasoning in high-dimensional observation spaces like images;
(2) diffusion action-heads provide flexibility in learning multi-modal action distributions, capturing multiple homotopy classes;
(3) the foundation model can leverage data from multiple robot embodiments of different sizes and traversability capabilities. However, System 1 is a black-box model, which cannot provide any safety guarantees.

\subsection{System 2: Physics-based Risk Estimation}
System 2 complements System 1 by providing interpretable safety guarantees for our specific robot and task. Like the thoughtful, deliberate System 2 described in \cite{daniel2017thinking}, our System 2 leverages inference-time compute to estimate the risk associated with given actions.

Specifically, we use a physics-based stochastic traversability estimate \cite{dixit2023step} to create risk maps from ego-centric $2.5D$ maps $m$. This estimate is based on a traversability cost $r = \mathcal{R}(m,x,u)$ representing the degree to which the robot can traverse a given state.
This becomes a random variable $R: (\mathcal{M}\times \mathcal{U}) \to \mathbb{R}$ via the belief $p(m|u,o)$. We employ CVaR as a risk metric:
\begin{align}
    \rho(R) = \text{CVaR}_\alpha(R) = \inf_{z\in \mathbb{R}} \mathbb{E} \left[ z + \frac{(R-z)_{+}}{1-\alpha} \right]
\end{align}
where $(.)_+ = max(.,0)$ and $\alpha \in (0,1]$ denotes the risk probability level.
Finally, we obtain a safe set of control inputs by obtaining a user-defined risk-tolerance threshold $\gamma$ as follows:
\begin{align}
\mathbbm{1}_{u \in U_{SAFE}}(u) =
\begin{cases}
   1 & \rho(R(u)) \le \gamma \\
   0 & otherwise
\end{cases}
\end{align}

\subsection{Thinking fast and slow: combining System 1 and 2}
We develop a unified mathematical approach that integrates both systems for safe and adaptive navigation. Mathematically, this can be formulated as a constrained sampling problem:
\begin{equation}
    u \sim p(u | z) = \pi(u | z) \mathbbm{1}_{u \in U_{SAFE}}(u),
    \label{eq:safe_distribution}
\end{equation}
where $\pi$ is the unconstrained policy of System 1's latent diffusion model.

A straight-forward approach is to use Classifier-like Risk Guidance \cite{dhariwal2021diffusion} by converting $\mathbbm{1}_{u \in U_{SAFE}}(u)$ to differentiable $c_{risk}(u)$ where $c_{risk}(u) \leq 0 \iff u \in U_{SAFE}$. This enables sampling from $u \sim p_{t,\theta}(u|z) e^{-\eta{c_{risk}(u)}}$ by modifying the diffusion score:
\begin{gather}
    \epsilon = \underbrace{\epsilon_\theta(u,z,t)}_{\text{foundation model}} + \underbrace{\eta\nabla_u c_{risk} (u)}_{\text{risk guidance}}.
\end{gather}

However, this penalty approach cannot guarantee constraint satisfaction since the implicit task objective $\min J(u,z)$ in $\pi(u|z) \propto \exp(-J(u,z))$ trades off with risk penalty.
Parameter sensitivity is severe: low $\eta$ violates constraints, while high $\eta$ eliminates valid modes (Fig.~\ref{fig:1d_example}).

\begin{algorithm}
\caption{Risk-Guidance Algorithm}
\begin{algorithmic}[1]
\Require Current sample $u^{t}$, risk map $\mathbbm{1}_{u \in U_{SAFE}}(u)$
\If{$t > t_2$}
    \While{$c_{risk}(\hat{u}, z, t) > 0$}
        \Comment{Rejection sample}
        \State $w \sim \mathcal{N}(0, I)$
        \State $\hat{u} \gets \frac{1}{\sqrt{\alpha^t}} \left( u^t - \frac{1-\alpha^t}{\sqrt{1-\bar{\alpha}^t}} \epsilon_\theta(u^t, z, t \right) + \sigma^t w$
    \EndWhile
\ElsIf{$t > t_1$}
    \While{$c_{risk}(\hat{u}, z, t) > 0$}
        \State $\hat{u} \gets (1-\alpha)\hat{u} + \alpha u^t$
        \Comment{Previous projection}
    \EndWhile
\Else
    \While{$c_{risk}(\hat{u}, z, t) > 0$}
        \State $\hat{u} \gets (1-\alpha)\hat{u}$
        \Comment{Small-action projection}
    \EndWhile
\EndIf
\State \textbf{return} $u^{t-1}$
\end{algorithmic} \label{alg:projection_sampling}
\end{algorithm}

These limitations can be addressed with a projection-based guidance strategy \cite{christopher2024constrained}.
The procedure: before each diffusion step, check if the next action violates the risk constraint.
If so, project to a nearby safe action; otherwise continue diffusion.
This guarantees constraint satisfaction by keeping actions in the safe set.
This strategy is effective (Fig.~\ref{fig:1d_example}, last row).
Unlike naive safety filters, enforcing constraints during diffusion spreads probability from constraint boundaries, encouraging exploration in informative areas.
A key challenge in applying this strategy is that projecting actions to the safe set is non-unique and in general the robot motion planning problem is PSPACE-hard \cite{canny1988complexity, lavalle2006planning}.

We address these challenges by employing three domain-specific projection strategies, as outlined in Algorithm~\ref{alg:projection_sampling}.
\textbf{Rejection sampling:} draws new noise values $w$ until constraints are satisfied---this is inefficient for narrow passages but effective in early iterations.
\textbf{Previous projection:} assuming the robot started in the safe set and no actions enter unsafe set during the diffusion process, the previous action from the diffusion process will also be safe.
\textbf{Small action projection:} exploits the fact that the robot remains safe if stationary.

\section{Experiments \& Results}

\begin{figure*}
    \centering
    \includegraphics[width=\linewidth]{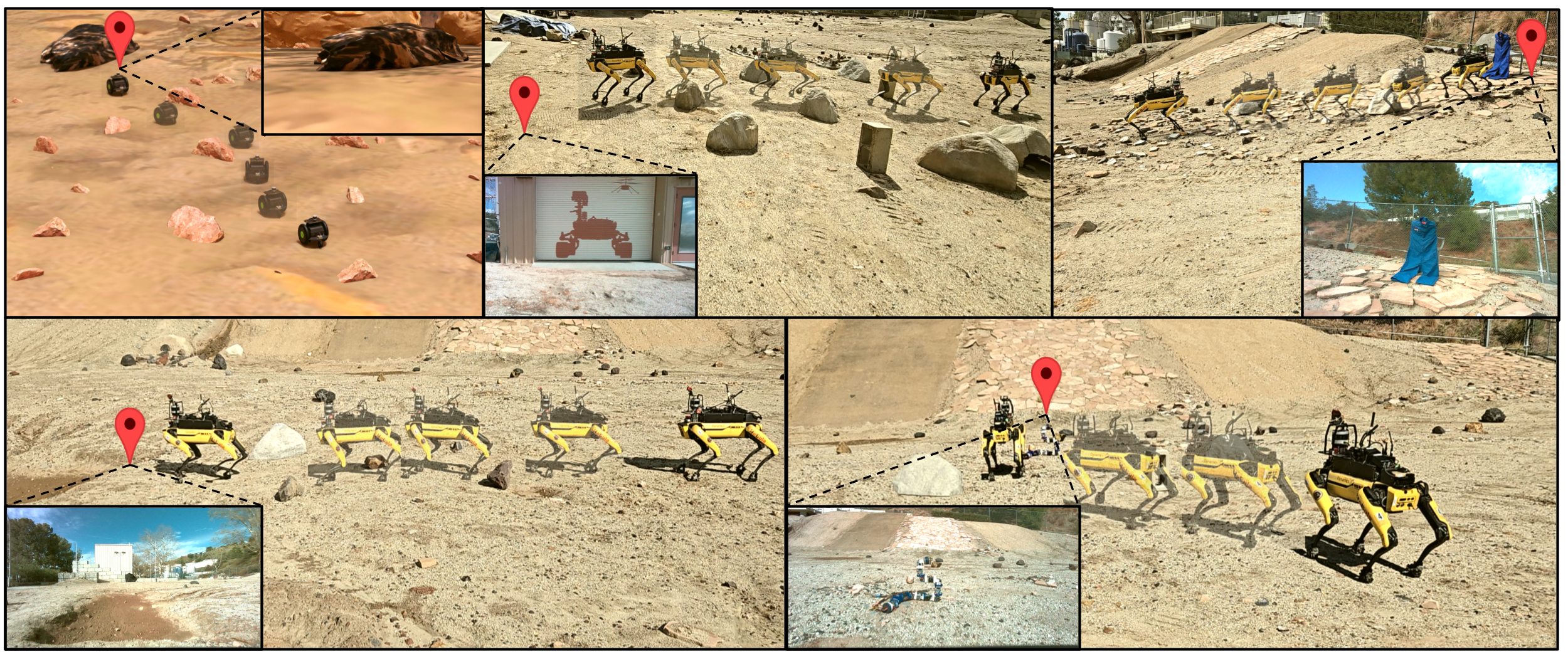}
    \caption{Test setups: simulation (top-left) and Mars Yars—Hard (top-right) \& Easy (bottom) with goal images.}
    \label{fig:exp_setup}
    \vspace{-3ex}
\end{figure*}

We conduct quantitative analysis in simulation using Isaac Sim with Carter robot and at the JPL Mars Yard using the NeBula Spot robot~\cite{bouman2020autonomous} with randomized start/goal configurations (Fig.~\ref{fig:exp_setup}).
We compare against: (1) Vanilla NoMaD \cite{sridhar2024nomad}, (2) Finetuned NoMaD, (3) NoMaD with safety filter, and (4) Risk-guided diffusion (Ours). Implementation details are provided in the appendix.
Our primary metrics are Goal Success rate (GS) and Safety Failure rate (SF). GS quantifies the percentage of trials where the robot successfully avoids obstacles and reaches the goal. SF quantifies the percentage of trials where the robot falls or steps on a high risk hazard.

\begin{table*}[t]
    \centering
    \begin{tabular}{cc}
        \begin{minipage}{0.40\textwidth}
            \centering
            \resizebox{\linewidth}{!}{
            \begin{threeparttable}
            \setlength{\tabcolsep}{0.5pt}
                \begin{tabular}{l@{\hspace{0.5cm}}cc}
                \toprule
                & \multicolumn{2}{c@{\hspace{0.5cm}}}{Mars Sim - Carter } \\
                \cmidrule(lr){2-3}
                \multirow{2}{*}{Approach} & Safety & Goal \\
                & Failure (\%) $\downarrow$ & Success (\%) $\uparrow$ \\
                \midrule \midrule
                Vanilla NoMaD & 100 & 0  \\
                \midrule
                NoMaD Finetuned & 50 & 46.67 \\
                \midrule
                NoMaD with Safety Filter & 0 & 46.67 \\
                \midrule
                \textbf{Our Method} & {\bfseries 0} & \textbf{50} \\
                \bottomrule
            \end{tabular}
            \caption{Simulation environment}
            \label{tab:quantitative_comparison_simulation}
            \begin{tablenotes}[flushleft]
            \centering
            \item
            \end{tablenotes}
           \end{threeparttable}}
        \end{minipage}
        &
        \begin{minipage}{0.59\textwidth}
            \centering
            \resizebox{\linewidth}{!}{
            \begin{threeparttable}
            \setlength{\tabcolsep}{0.5pt}
                \begin{tabular}{l@{\hspace{0.5cm}}cc@{\hspace{0.5cm}}cc}
                \toprule
                & \multicolumn{2}{c@{\hspace{0.5cm}}}{Mars Yard - Spot (Easy)} & \multicolumn{2}{c}{Mars Yard - Spot (Hard)} \\
                \cmidrule(lr){2-3} \cmidrule(lr){4-5}
                \multirow{2}{*}{Approach} & Safety & Goal & Safety & Goal \\
                & Failure (\%) $\downarrow$ & Success (\%) $\uparrow$ & Failure (\%) $\downarrow$ & Success (\%) $\uparrow$ \\
                \midrule \midrule
                Vanilla NoMaD & 100 & 0 & 100 & 0 \\
                \midrule
                NoMaD Finetuned & 30 & \textbf{70} & 48.57 & \textbf{25.92} \\
                \midrule
                \textbf{Our Method} & \textbf{6.67} & 66.67 & \textbf{14.28} & 23.07 \\
                \bottomrule
            \end{tabular}
            \caption{Real world environment}
            \label{tab:quantitative_comparison_hardware}
            \begin{tablenotes}[flushleft]
            \centering
            \item
            \end{tablenotes}
           \end{threeparttable}}
        \end{minipage}
    \end{tabular}
    \vspace{-4mm}
\end{table*}

Table~\ref{tab:quantitative_comparison_simulation} and \ref{tab:quantitative_comparison_hardware} show vanilla NoMaD leads to collisions in all of our simulation and real-world test cases. This shows that such models are very brittle when it comes to out-of-distribution scenarios. Even after fine-tuning, the model is not capable of avoiding collisions in a majority of the test cases. In contrast, the safety filter and our method reduced the safety violations to $0$ in simulation and by almost $4 \times$ for real world environments.

Contrary to expectations, our method only slightly outperforms the safety filter.
The model’s bias toward out-of-distribution obstacles—and the training data’s failure to learn multimodal actions across homotopies—renders guidance no better than filtering.
In cluttered scenes the robot dodges obstacles but then loses sight of the goal; with limited memory it fails to re-acquire it, lowering success. Enhancing policy memory is a key next step.
We expect these limitations to reduce as foundation models grow in scale and improve architecturally, our modular risk-guided diffusion framework should inherit these gains with minimal changes.

\section{Conclusion and Future Work}
Risk-Guided Diffusion shows that robot navigation foundation models can be made mission-safe while retaining their semantic navigation capabilities.
A simple projection step—essentially a fast collision check rather than the quadratic programs typical of CBF pipelines—lets a learned diffusion policy collaborate with a physics-based risk map, cutting safety violations up to $ 4\times$ while maintaining goal success at the JPL Mars Yard.

The current gains over a basic safety filter are modest, limited by trajectory diversity and short-term memory in today’s foundation models. We therefore invite the community to push these fronts—richer multimodal training, longer-horizon memory, and tighter guarantees—so that the method can mature into a dependable navigator for Mars lava tubes, the icy terrains of Europa and Enceladus, and other uncharted worlds.

\bibliographystyle{IEEEtran}
\bibliography{ref}

\begin{thebibliography}{10}
\providecommand{\url}[1]{#1}
\csname url@samestyle\endcsname
\providecommand{\newblock}{\relax}
\providecommand{\bibinfo}[2]{#2}
\providecommand{\BIBentrySTDinterwordspacing}{\spaceskip=0pt\relax}
\providecommand{\BIBentryALTinterwordstretchfactor}{4}
\providecommand{\BIBentryALTinterwordspacing}{\spaceskip=\fontdimen2\font plus
\BIBentryALTinterwordstretchfactor\fontdimen3\font minus
  \fontdimen4\font\relax}
\providecommand{\BIBforeignlanguage}[2]{{%
\expandafter\ifx\csname l@#1\endcsname\relax
\typeout{** WARNING: IEEEtran.bst: No hyphenation pattern has been}%
\typeout{** loaded for the language `#1'. Using the pattern for}%
\typeout{** the default language instead.}%
\else
\language=\csname l@#1\endcsname
\fi
#2}}
\providecommand{\BIBdecl}{\relax}
\BIBdecl

\bibitem{daniel2017thinking}
K.~Daniel, \emph{Thinking, fast and slow}, 2017.

\bibitem{ono2024boldly}
M.~Ono \emph{et~al.}, ``To boldly go where no robots have gone before--part 1:
  Eels robot to spearhead a new one-shot exploration paradigm with in-situ
  adaptation,'' in \emph{AIAA Scitech Forum}, 2024.

\bibitem{thakker2024boldly}
R.~Thakker \emph{et~al.}, ``To boldly go where no robots have gone before--part
  4: Neo autonomy for robustly exploring unknown, extreme environments with
  versatile robots,'' in \emph{AIAA SCITECH Forum}, 2024.

\bibitem{kim2024openvla}
M.~J. Kim, K.~Pertsch, S.~Karamcheti, T.~Xiao, A.~Balakrishna, S.~Nair,
  R.~Rafailov, E.~Foster, G.~Lam, P.~Sanketi \emph{et~al.}, ``Openvla: An
  open-source vision-language-action model,'' \emph{arXiv preprint
  arXiv:2406.09246}, 2024.

\bibitem{collaboration2023open}
O.-X.~E. Collaboration, A.~Padalkar, A.~Pooley, A.~Jain, A.~Bewley, A.~Herzog,
  A.~Irpan, A.~Khazatsky, A.~Rai, A.~Singh \emph{et~al.}, ``Open x-embodiment:
  Robotic learning datasets and rt-x models,'' \emph{arXiv preprint
  arXiv:2310.08864}, 2023.

\bibitem{shah2023gnm}
D.~Shah, A.~Sridhar, A.~Bhorkar, N.~Hirose, and S.~Levine, ``Gnm: A general
  navigation model to drive any robot,'' in \emph{2023 IEEE International
  Conference on Robotics and Automation (ICRA)}.\hskip 1em plus 0.5em minus
  0.4em\relax IEEE, 2023, pp. 7226--7233.

\bibitem{dixit2023step}
A.~Dixit, D.~D. Fan, K.~Otsu, S.~Dey, A.-A. Agha-Mohammadi, and J.~W. Burdick,
  ``Step: Stochastic traversability evaluation and planning for risk-aware
  off-road navigation; results from the darpa subterranean challenge,''
  \emph{arXiv preprint arXiv:2303.01614}, 2023.

\bibitem{freyfast}
J.~Frey, M.~Mattamala, N.~Chebrolu, C.~Cadena, M.~Fallon, and M.~Hutter,
  ``{Fast Traversability Estimation for Wild Visual Navigation},'' in
  \emph{Proceedings of Robotics: Science and Systems}, Daegu, Republic of
  Korea, July 2023.

\bibitem{setlur2025scaling}
A.~Setlur, N.~Rajaraman, S.~Levine, and A.~Kumar, ``Scaling test-time compute
  without verification or rl is suboptimal,'' \emph{arXiv preprint
  arXiv:2502.12118}, 2025.

\bibitem{Liang2024DTGD}
J.~Liang, A.~Payandeh, D.~Song, X.~Xiao, and D.~Manocha, ``Dtg :
  Diffusion-based trajectory generation for mapless global navigation,''
  \emph{IEEE International Conference on Intelligent Robots and Systems
  (IROS)}, 2024.

\bibitem{xiao2023safediffuser}
W.~Xiao, T.-H. Wang, C.~Gan, and D.~Rus, ``Safediffuser: Safe planning with
  diffusion probabilistic models,'' \emph{arXiv preprint arXiv:2306.00148},
  2023.

\bibitem{ames2016control}
A.~D. Ames, X.~Xu, J.~W. Grizzle, and P.~Tabuada, ``Control barrier function
  based quadratic programs for safety critical systems,'' \emph{IEEE
  Transactions on Automatic Control}, vol.~62, no.~8, pp. 3861--3876, 2016.

\bibitem{bjorck2025gr00t}
J.~Bjorck, F.~Casta{\~n}eda, N.~Cherniadev, X.~Da, R.~Ding, L.~Fan, Y.~Fang,
  D.~Fox, F.~Hu, S.~Huang \emph{et~al.}, ``Gr00t n1: An open foundation model
  for generalist humanoid robots,'' \emph{arXiv preprint arXiv:2503.14734},
  2025.

\bibitem{figure2024helix}
A.~Figure, ``Helix: A vision-language-action model for generalist humanoid
  control,'' \emph{Figure AI News}, 2024.

\bibitem{yoon2025monte}
J.~Yoon, H.~Cho, D.~Baek, Y.~Bengio, and S.~Ahn, ``Monte carlo tree diffusion
  for system 2 planning,'' \emph{arXiv preprint arXiv:2502.07202}, 2025.

\bibitem{sridhar2024nomad}
A.~Sridhar, D.~Shah, C.~Glossop, and S.~Levine, ``Nomad: Goal masked diffusion
  policies for navigation and exploration,'' in \emph{IEEE International
  Conference on Robotics and Automation (ICRA)}, 2024.

\bibitem{chi2024diffusionpolicy}
C.~Chi, Z.~Xu, S.~Feng, E.~Cousineau, Y.~Du, B.~Burchfiel, R.~Tedrake, and
  S.~Song, ``Diffusion policy: Visuomotor policy learning via action
  diffusion,'' \emph{The International Journal of Robotics Research}, 2024.

\bibitem{dhariwal2021diffusion}
P.~Dhariwal and A.~Nichol, ``Diffusion models beat gans on image synthesis,''
  \emph{Advances in neural information processing systems}, vol.~34, pp.
  8780--8794, 2021.

\bibitem{christopher2024constrained}
J.~K. Christopher, S.~Baek, and N.~Fioretto, ``Constrained synthesis with
  projected diffusion models,'' \emph{Advances in Neural Information Processing
  Systems}, 2024.

\bibitem{song2020score}
Y.~Song, J.~Sohl-Dickstein, D.~P. Kingma, A.~Kumar, S.~Ermon, and B.~Poole,
  ``Score-based generative modeling through stochastic differential
  equations,'' in \emph{International Conference on Learning Representations
  (ICLR)}, 2021.

\bibitem{canny1988complexity}
J.~Canny, ``The complexity of robot motion planning,'' \emph{MIT press}, 1988.

\bibitem{lavalle2006planning}
S.~M. LaValle, \emph{Planning algorithms}.\hskip 1em plus 0.5em minus
  0.4em\relax Cambridge university press, 2006.

\bibitem{bouman2020autonomous}
A.~Bouman \emph{et~al.}, ``Autonomous spot: Long-range autonomous exploration
  of extreme environments with legged locomotion,'' in \emph{IEEE International
  Conference on Intelligent Robots and Systems (IROS)}, 2020.

\end{thebibliography}

\begin{appendices}

\section{Baseline implementation details}
\begin{enumerate}
    \item Vanilla NoMaD : For the Vanilla NoMaD baseline, we utilize the publicly available pre-trained weights released by the original authors and evaluate the model in a zero-shot setting across both simulated and real-world environments. We adhere to the normalization parameters provided in the NoMaD framework and apply velocity-based unnormalization to obtain the final action outputs from the diffusion model.
    \item Finetuned NoMaD : We collect approximately one hour of driving data in both simulated and real-world environments via teleoperation of the robot, following the methodology employed in the open-source datasets used by the NoMaD framework. For data preprocessing, we adopt a distance-based waypoint sampling strategy applied across both our collected datasets and the original NoMaD datasets. Waypoints are sampled at uniform intervals of 0.2 meters, or at shorter distances when the change in heading exceeds a predefined angular threshold. Using this dataset, we first train the original NoMaD model using the same hyperparameters as the baseline, and subsequently fine-tune the model on our custom dataset.
    \item Safety Filter: To ensure safety, we implement a simple truncation-based safety filter that truncates the output trajectory at the waypoint immediately preceding the first predicted collision. This approach guarantees that the resulting trajectory remains entirely within safe bounds.
    \item Risk Guidance Diffusion : We implement the projected risk guidance mechanism as described in this work, utilizing the risk map generated using \cite{dixit2023step}. In simulation, we sample a total of 50 trajectories, while in the real-world setting, we sample 8 trajectories. Risk-guidance is performed in parallel across all trajectories to ensure fast inference.
\end{enumerate}

\end{appendices}

\end{document}